\documentclass[conference]{IEEEtran}
\IEEEoverridecommandlockouts
\usepackage{cite}
\usepackage{amsmath,amssymb,amsfonts}
\usepackage{algorithmic}
\usepackage{graphicx}
\usepackage{textcomp}
\usepackage{multicol}
\usepackage{xcolor}
\usepackage{float}
\usepackage{subcaption,booktabs}

\usepackage{algorithm2e}
\usepackage[english]{babel}

\def\BibTeX{{\rm B\kern-.05em{\sc i\kern-.025em b}\kern-.08em
    T\kern-.1667em\lower.7ex\hbox{E}\kern-.125emX}}
    
\usepackage{todonotes}
\newcommand*{\red}{\textcolor{red}}

\begin{document}

\title{BiTrackGAN: Cascaded CycleGANs to Constraint Face Aging \\
}

\author{\IEEEauthorblockN{
Tsung-Han Kuo\IEEEauthorrefmark{1},
Zhenge Jia\IEEEauthorrefmark{2}, 
Tei-Wei Kuo\IEEEauthorrefmark{1}, and
Jingtong Hu\IEEEauthorrefmark{2},
}

\IEEEauthorblockA{
\IEEEauthorrefmark{1}
Graduate Institute of Networking and Multimedia, National Taiwan University \\
\IEEEauthorrefmark{2}Department of Electrical and Computer Engineering, University of Pittsburgh \\
Email: 
\IEEEauthorrefmark{1}d05944019@ntu.edu.tw,
\IEEEauthorrefmark{2}zhenge.jia@pitt.edu,
\IEEEauthorrefmark{1}ktw@csie.ntu.edu.tw,
\IEEEauthorrefmark{2}jthu@pitt.edu
}
}

\maketitle

\begin{abstract}
With the increased accuracy of modern computer vision technology, many access control systems are equipped with face recognition functions for faster identification. In order to maintain high recognition accuracy, it is necessary to keep the face database up-to-date. However, it is impractical to collect the latest facial picture of the system's user through human effort. Thus, we propose a bottom-up training method for our proposed network to address this challenge. Essentially, our proposed network is a translation pipeline that cascades two CycleGAN blocks (a widely used unpaired image-to-image translation generative adversarial network) called BiTrackGAN. By bottom-up training, it induces an ideal intermediate state between these two CycleGAN blocks, namely the constraint mechanism. Experimental results show that BiTrackGAN achieves more reasonable and diverse cross-age facial synthesis than other CycleGAN-related methods. As far as we know, it is a novel and effective constraint mechanism for more reason and accurate aging synthesis through the CycleGAN approach. 
\end{abstract}

\begin{IEEEkeywords}
Facial Aging, CycleGAN, Mode Collapse, Constraint Mechanism, Bottom-Up Training
\end{IEEEkeywords}

\section{Introduction}
\label{I_intro}

Since deep learning took off in 2012, computer vision has improved significantly in recognition and detection. There are so many access control systems that come equipped with a face recognition function for faster identification. 
In order to maintain high accuracy, facial recognition relies heavily on the newest facial image database for training models and comparisons. However, it is impractical to collect the latest facial picture of the system's user through human effort.  Therefore, a novel cross-age face synthesis technology is required to keep the face database up-to-date, particularly in regard to unpaired image-to-image translation.

Current cross-age face images synthesis methods can be classified into three categories: prototype-based synthesis~\cite{kemelmacher2014illumination}, physical model-based synthesis~\cite{suo2012concatenational, suo2009compositional}, and deep learning-based synthesis~\cite{huang2020pfa, fang2020triple, liu2017face}. 
Prototype and physical model-based synthesis methods usually require extensive face images to construct the prototype or highly rely on the manually extracted features (e.g., eyebrow, hair color, skin color, wrinkle, etc). 
On the contrary, deep learning-based synthesis methods could automatically learn to synthesize high fidelity natural face images by invoking generative adversarial network (GAN). 
\cite{liu2017face} proposes Contextual Generative Adversarial Nets (C-GANs) that learns to generate face images with input ages as a condition. Authors in~\cite{fang2020triple} propose a novel Tripe Generative Adversarial Networks (Triple-GAN) by devising a triple translation loss to better simulate the face aging changes. Authors in~\cite{huang2020pfa} develop a progressive face aging generative adversarial network (PFA-GAN) that contains sub-networks to learn aging effects between two age groups and mimic the face aging procedure.

Although current deep learning-based synthesis methods could generate high fidelity natural face images via GAN, there are still several challenges when applied in practical scenarios:

1) Deep learning-based methods heavily rely on a large number of annotated and paired cross-age facial image samples for training. The acquisition of a such training dataset is hard to obtain in practical scenarios.

2) GAN-based models cannot prevent mode collapse. The authors in~\cite{arora2017generalization} have proven that, even though the generated distribution is far from target distribution, the training objective can approach optimum value.

3) In CycleGAN, the translation effect is weaker when the difference between the two domains is slight. Experiments in ~\cite{palsson2018generative} have demonstrated this.

To address these challenges, we first proposed a constraint mechanism for the CycleGAN approach. Specifically, the constraint mechanism is the ideal intermediate state of cascading multiple progressive translation pipelines. These ideal intermediate states will constrain the entire translation pipeline to output a smoother, progressive translated synthesis.



Furthermore, we suggest a bottom-up training manner for our proposed network. Essentially, our proposed network is a translation pipeline that cascades two CycleGAN blocks, called BiTrackGAN. In BiTrackGAN, different generators with the same objective, sharing the same network architecture, modify each other's optimal parameters during the optimization. This way, other generators will be forced to re-explore better distributions to approach the target and produce a more ideal intermediate state. That's how bottom-up training induces the ideal intermediate state, the constraint mechanism.

The experiment results show that BiTrackGAN synthesizes smoother and progressive facial aging and rejuvenation with a constraint mechanism. Furthermore, it improves synthesis diversity.

The main contributions in this paper are summarized as follows:
\begin{itemize}
    \item We proposed a constraint mechanism for the CycleGAN approach to improve facial aging and rejuvenation synthesis.
    
    \item We propose a bottom-up training method for a larger network that is a translation pipeline with two CycleGAN blocks cascaded to synthesize facial aging and rejuvenation, named BiTrackGAN. By bottom-up training, it induces an ideal intermediate state between these two CycleGAN blocks, namely the constraint mechanism.
\end{itemize}


The rest of the paper summarized as follows: 
Section~\ref{II_related} presents the related works.
Section~\ref{III_method} introduces the methodology. 
Section~\ref{IV_experiments} demonstrates the experimental setup and results. 
Section~\ref{V_Conclusion} concludes the paper. 

\section{Related Work}
\label{II_related}

\subsection{Conventional Face Aging Approaches}

There two main categories in conventional face aging approaches: physical model-based synthesis and prototype-based synthesis.

Physical model-based synthesis approaches aim to explicitly describe the appearance of an individual cross-age by establishing a parameterized model. 
Authors in~\cite{suo2009compositional} propose a dynamic model which contains a hierarchical And-Or graph to describe different parts of faces cross-age. 
Authors in~\cite{suo2012concatenational} devise a concatenational graph that models the long-term face aging process.
However, physical model-based approaches require an extensive amount of training samples (i.e., cross-age face images) to develop the model. 

Prototype-based synthesis approaches focus on constructing a prototype face image for each age group. 
Authors in~\cite{kemelmacher2014illumination} construct a prototype face image in each  age group from thousands of collected face images of children and adults.
For each testing individual, the prototype face image be further fine-tuned by reflecting the differences in shape and texture onto the image. 
Authors in~\cite{tiddeman2001prototyping} utilizes the prototype face images to determine the transferring features between two different prototype faces. 
The defined transferring features would be then applied to conduct face age synthesis with the given input face image. 
However, the main drawback of the prototype-based could ignore the personal information of testing individual and further degrade the quality of the synthesized face images. 

\subsection{Deep learning Based Face Aging}

Generative adversarial network (GAN) has been widely adopted in face age synthesis applications. 
GAN usually consists of two essential components: Generator $G$ that generates data and Discriminator $D$ that discriminates data to be real or generated.
$G$ and $D$ would be trained in an adversarial manner where $G$ would generate the data as real as possible to deceive $D$ while $D$ would be trained to discriminate the generated data from other real input data. 

Authors in~\cite{liu2017face} propose Contextual Generative Adversarial Nets (C-GANs) that utilize ages as a condition to enable GAN to synthesize face images imitating the aging procedure. 

Authors in~\cite{fang2020triple} propose a novel Tripe Generative Adversarial Networks (Triple-GAN) by devising a triple translation loss to better simulate the face aging changes. 

Authors in~\cite{huang2020pfa} develop a progressive face aging generative adversarial network (PFA-GAN), which contains several sub-networks learning aging effects between two age groups.  

In~\cite{palsson2018generative}, authors proposes Group-GAN, a group of CycleGANs trained for face aging and rejuvenation synthesis for adjacent age groups. However, the synthesis effect is weak when the difference between the two groups is slight.
 

\subsection{Mode Collapse Evaluation}
In GAN approach, mode collapse cannot be prevented. Thus, we need a metric for measuring the degree of mode collapse in a GAN model. The NDB Score is a metric that evaluates mode collapse; it was proposed as part of the work~\cite{richardson2018gans}. The metric belongs to the quantitative analysis approach that is efficient to identify mode collapse without going over all synthesis images or training loss graphs. 

Based on the work~\cite{richardson2018gans}, the following steps are roughly summed up in calculating the NDB Score:

1) Divide the training set into K clusters using a clustering algorithm such as K-Means.

2) Allocating synthesis samples to those K clusters based on the Euclidean distance between sample points and centroids of K clusters.

3) Calculating Z-Score by testing synthesis samples and training samples for each cluster, then marking one cluster as statistically different if the Z-Score is less than 0.5.

All of these statistically different clusters can now be referred to as statistically different bins, namely the Number of statistically Different Bins (NDB).  By dividing NDB by K, we can normalize it between 0 and 1, with a lower NDB/K value indicating a lower mode collapse degree.
\section{Method}
\label{III_method}

In this section, we will first introduce the core component CycleGAN block in BiTrackGAN, then show how to train the entire translation pipeline in a bottom-up manner.

\subsection{Network Architecture}

\begin{figure*}[t]
\includegraphics[width = 1.0\textwidth]{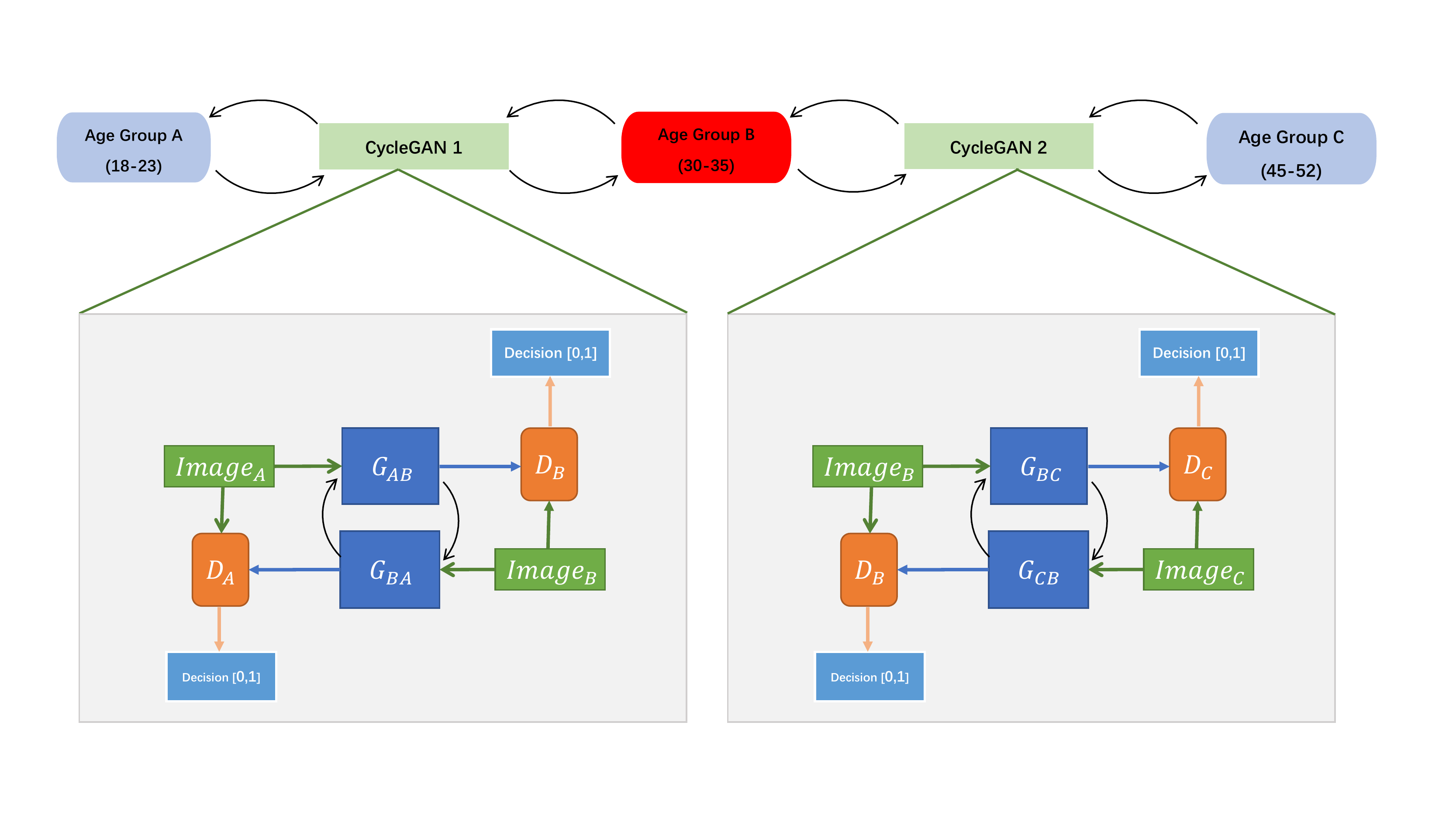}
\centering
\caption{
BiTrackGAN Architecture. Essentially, it's a translation pipeline cascaded by two CycleGAN blocks. The intermediate state (red blocks), the constraint mechanism, constrains the entire translation pipeline to produce a smoother, more progressive translation synthesis.
}
\label{fig:bitrackgan}
\end{figure*}

As shown in Figure~\ref{fig:bitrackgan}, BiTrackGAN is a translation pipeline cascaded by two CycleGAN blocks. For cross-age facial image synthesis, CycleGAN 1 first translates the source image into an intermediate state image that represents age group B. CycleGAN 2 then translates that to a final synthesis facial image for group C, and vice versa. Considering that CycleGAN is a core component of BiTrackGAN, we will briefly introduce its structure and training approach.

\subsection{CycleGAN}
CycleGAN is a great work that addresses unpaired image-to-image translation in~\cite{zhu2017unpaired}. The main idea is to build a relationship between the source image and generated image by leveraging cycle consistency. In this way, it enables image-to-image translation without paired data. 
CycleGAN requires two generators, one to translate images from the source domain into the target domain, and another to translate back from the target domain into the source domain. Additionally, two discriminators are required during the training phase to determine whether the translated image reaches the target domain. CycleGAN has the same internal structure as the CycleGAN blocks shown in the Fig.~\ref{fig:bitrackgan}.
In CycleGAN 1,  forward cycle translation consists of $G_{AB}$ and $G_{BA}$. This can be viewed as an adversarial autoencoder, in which the bottleneck is trained with an adversarial loss to achieve the target domain. The encoder part translates an image from the source domain to the target domain, and the decoder part reconstructs it from the target domain to the source domain. Similarly, backward cycle translation works in the same way. In this way, we establish the relationship between the source image and generated image, and we can address the unpaired image problem.

\begin{algorithm}
\caption{Bottom-Up Training}
\label{alg:bottom-up-training}
\SetKwInput{KwInput}{Input}                 
\SetKwInput{KwOutput}{Output}              

\DontPrintSemicolon
  
  \KwInput{$ G_{AB}, G_{BC}, D_{A}, D_{B}, D_{C}$}
  \KwData{Dataset $S_A$, $S_B$, $S_C$}
  \;
  
  \SetKwFunction{FMain}{BottomUpTraining}
  \SetKwFunction{FTrainDiscriminators}{TrainDis}
  \SetKwFunction{FTrainGenerators}{TrainGen}
  \SetKwFunction{FTrainCycleGAN}{TrainCG}

  \SetKwProg{Fn}{Def}{:}{}
  \tcc{Discriminators Training}
  \Fn{\FTrainDiscriminators{$ G_{XY}, G_{YX}, D_{X}, D_{Y}, s_x, s_y $}}{
        $\mathcal{L}_{D_Y}(G_{XY}, D_{Y}, S_X, S_Y) = $ 
        $\mathbb{E}_{s_y \sim p_{data}(s_y)}[(D_{Y}(s_y)-1)^2] + $
        $\mathbb{E}_{s_x \sim p_{data}(s_x)}[D_{Y}(G_{XY}(s_a))^2]$ \\
        \;
        
        $\mathcal{L}_{D_X}(G_{YX}, D_{X}, S_X, S_Y)$ = 
        $\mathbb{E}_{s_x \sim p_{data}(s_x)}[(D_{X}(s_x)-1)^2]$ + 
        $\mathbb{E}_{s_y \sim p_{data}(s_y)}[D_{X}(G_{YX}(s_y))^2]$ \\
        \;
        
        $\min \mathcal{L}_{D} = \mathcal{L}_{D_Y} + \mathcal{L}_{D_X}$ 
        
  }
  \;

  \SetKwProg{Fn}{Def}{:}{}
  \tcc{Generators Training}
  \Fn{\FTrainGenerators{$ G_{XY}, G_{YX}, D_{X}, D_{Y}, s_x, s_y $}}{
        
        $\mathcal{L}_{G_{XY}}(G_{XY}, D_{Y}, S_X, S_Y) = $ 
        $\mathbb{E}_{s_x \sim p_{data}(s_x)}[(D_{Y}(G_{XY}(s_x))-1)^2] + $
        $\mathcal{L}_{cycle_{X-Y-X}} $\\
        \;
        
        $\mathcal{L}_{G_{YX}}(G_{YX}, D_{X}, S_Y, S_X) = $
        $\mathbb{E}_{s_y \sim p_{data}(s_y)}[(D_{X}(G_{YX}(s_y))-1)^2] + $
        $\mathcal{L}_{cycle_{Y-X-Y}} $\\
        \;
        
        $\min \mathcal{L}_{G} = \mathcal{L}_{G_{XY}} + \mathcal{L}_{G_{YX}} $
        
  }
  \;
  
  \SetKwProg{Fn}{Def}{:}{}
  \tcc{CycleGAN Training}
  \Fn{\FTrainCycleGAN{$ G_{XY}, G_{YX}, D_{X}, D_{Y}, s_x, s_y $}}{
  
        TrainDis($ G_{XY}, G_{YX}, D_{X}, D_{Y}, s_x, s_y $)\\
        TrainGen($ G_{XY}, G_{YX}, D_{X}, D_{Y}, s_x, s_y $)
        
  }
  \;

  \SetKwProg{Fn}{Def}{:}{}
  \SetKw{KwBy}{by}
  \Fn{\FMain{$ G_{AB}, G_{BC}, D_{A}, D_{B}, D_{C}$}}{
        $G_{ABC} = G_{BC} \circ G_{AB}$\\
        $G_{CBA} = G_{BA} \circ G_{CB}$\\
        \;
        \tcc{For each epoch loop}
        \For{$i\gets1$ \KwTo $200$ \KwBy $1$}{
            \tcc{For each iteration loop}
            \For{$j\gets1$ \KwTo $200$ \KwBy $1$}{
                 \tcc{Draw minibatch samples}
                $s_a^j$ $\in$ $S_A$, $s_b^j$ $\in$ $S_B$, $s_c^j$ $\in$ $S_C$ \\
                \;
                \tcc{Train CycleGAN 1}
                TrainCG($ G_{AB},   G_{BA},   D_{A}, D_{B}, s_a^j, s_b^j $)\\
                \;
                \tcc{Train CycleGAN 2}
                TrainCG($ G_{BC},   G_{CB},   D_{B}, D_{C}, s_b^j, s_c^j $)\\
                \;
                \tcc{Train BiTrackGAN}
                TrainCG($ G_{ABC}, G_{CBA}, D_{A}, D_{C}, s_a^j, s_c^j $)
            }
        }
  }
\end{algorithm}
Formally, CycleGAN optimization is also a min-max problem. In order to achieve stable training in practice, the author in~\cite{zhu2017unpaired} changed the objective loss to the least-squares loss. Therefore, the loss function for optimizing discriminators and generators is the same as for TrainDis and TrainGen methods in Algorithm~\ref{alg:bottom-up-training}. Moreover, for each iteration, the generators will be frozen first, followed by the discriminator optimization; then the discriminator will be frozen, followed by the generators optimization. In this way, CycleGAN training is similar to TrainCG method in Algorithm~\ref{alg:bottom-up-training}.

\subsection{Bottom-Up Training}
Eventually, we will need to train the entire translation pipeline, namely BiTrackGAN. As mentioned earlier, it contains two CycleGAN blocks; to induce an ideal intermediate state as red block shown in Figure~\ref{fig:bitrackgan}, we use a bottom-up gradual training approach. In other words, for each iteration, CycleGAN 1 is trained first, followed by CycleGAN 2, then the whole BiTrackGAN is trained, as in BottomUpTraining method in Algorithm~\ref{alg:bottom-up-training}. 

\subsection{Constraint Mechanism}
By bottom-up training, it induces an ideal intermediate state between these two CycleGAN blocks, namely the constraint mechanism. As mentioned early, the BiTrackGAN is a translation pipeline that cascades two CycleGAN blocks, different generators with the same objective, sharing the same network architecture, modify each other's optimal parameters during the self-optimization. Therefore, other generators will have to re-explore better distributions in order to approach the target and produce a better intermediate state. As a result, bottom-up training induces the ideal intermediate state, namely constraint mechanisms.

In summary, we propose a bottom-up training method for a network that is a translation pipeline cascaded two CycleGAN blocks to synthesize facial aging and rejuvenation, named BiTrackGAN. By bottom-up training, it induces an ideal intermediate state between these two CycleGAN blocks, namely the constraint mechanism. As a result, BiTrackGAN produces a smoother and more progressive facial aging and rejuvenation, as well as improved synthesis diversity.
\section{Experiments}
\label{IV_experiments}

In this section, we first present the experimental setup and implementation details. 
We then demonstrate experimental results together with the discussions of the proposed methods. 

\subsection{System Setup}

We conduct training on an workstation which is equipped with 1 Ampere's Altra 80 Cores ARM Processor running at 3.0 GHz, 502G memory, and 2 NVIDIA A100 GPUs with 40G memory. We implemented BiTrackGAN on the deep learning framework pytorch (1.9.0), and Python (3.8.10).

\subsection{Data Preparation}

We utilize the MORPH-II Longitudinal Database~\cite{ricanek2006morph} for training and evaluation in our experiments. 
In our experiment, we have three age groups: 20s (18-23), 30s (30-35), and 50s (45-52). For each age group, 4,000 face images are used for training and 200 for testing. To avoid data leakage during training, each face image would only be included in training or testing set. 

As far as we know, different races or occupations will have different degrees of aging effect, which is a well-known challenge. Using the annotation file, the dataset is divided into white and black races, and only black race face images are used to train and evaluate the model.

\subsection{Implementation Details}

The CycleGAN blocks utilized in BiTrackGAN are cascaded following the descriptions in Section~\ref{III_method}. 
In each CycleGAN block, we just leverage the original CycleGAN architecture introduced in~\cite{zhu2017unpaired}, which includes some down-sampling Conv layers, Transformers (9 ResNet blocks) at the bottleneck, and up-sampling deconvolution layers.

During training, we adopted the Adam as an optimizer with a learning rate=0.0002, and the exponential decay rate is 0.5 for the 1st-moment estimates. The number of training epochs is 200, and the batch size is 20 for each age group. 

In the bottom-up training, the first CycleGAN block is trained in each iteration, followed by the second and the entire BiTrackGAN as Algorithm~\ref{alg:bottom-up-training}. In this way, BiTrackGAN achieved a more smooth, progressively aging effect and a greater degree of diversity. The total duration of all the training on our system is about 26 hours.

\begin{figure*}[htbp]
    \centering
        \begin{tabular}{cccccccccc}
            \multicolumn{9}{c}{$\overbrace{\rule{41.5em}{0pt}}^{\text{Face Aging}}$}\\
            Age 18-23 && \multicolumn{2}{c}{Age 30-35}     && \multicolumn{2}{c}{Age 45-52}     && Age 45-52\\
            Medium=20 && \multicolumn{2}{c}{Medium=33}     && \multicolumn{2}{c}{Median=48}     && Median=48\\

            \includegraphics[width=0.13\textwidth]{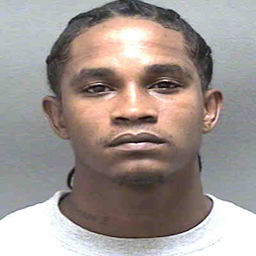} &&
            \includegraphics[width=0.13\textwidth]{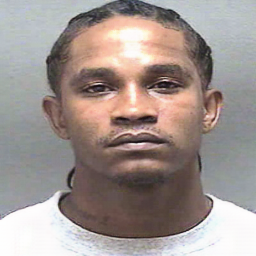} &
            \includegraphics[width=0.13\textwidth]{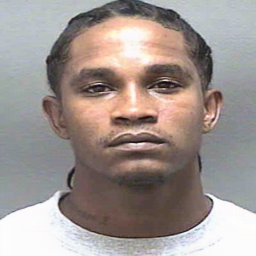} &&
            \includegraphics[width=0.13\textwidth]{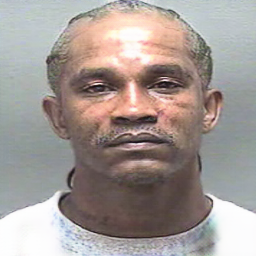} &
            \includegraphics[width=0.13\textwidth]{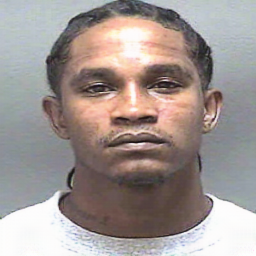} &&
            \includegraphics[width=0.13\textwidth]{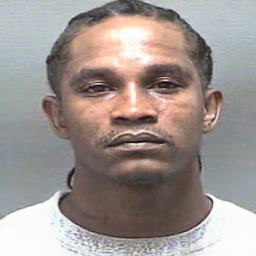} \\
                26  && \red{33 (+0)} & 29 (-4) && \red{51 (+3)} & 37 (-11) && 39 (-9) \\
            
            \includegraphics[width=0.13\textwidth]{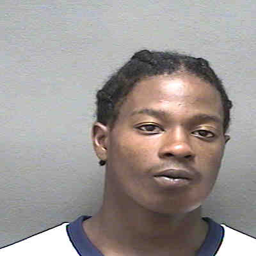} &&
            \includegraphics[width=0.13\textwidth]{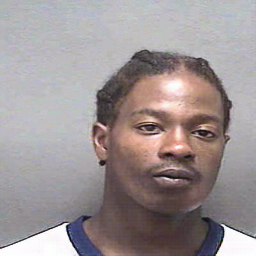} &
            \includegraphics[width=0.13\textwidth]{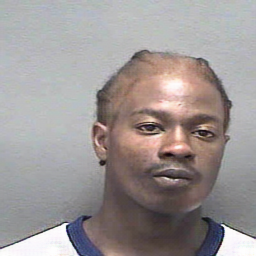} &&
            \includegraphics[width=0.13\textwidth]{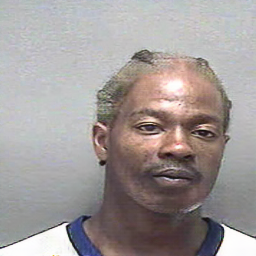} &
            \includegraphics[width=0.13\textwidth]{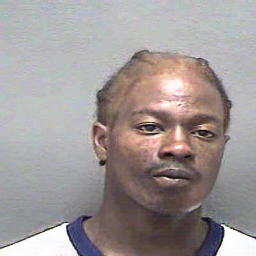} &&
            \includegraphics[width=0.13\textwidth]{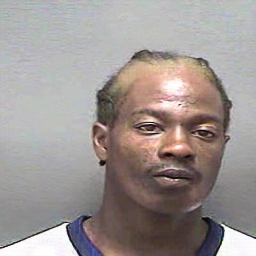} \\
                25  && \red{29 (-4)} & 34 (+1) && \red{48 (+0)} & 45 (-3)  && 45 (-3) \\
            
            \includegraphics[width=0.13\textwidth]{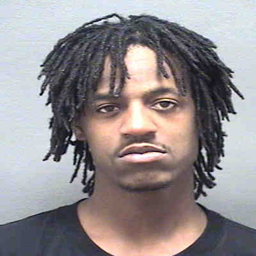} &&
            \includegraphics[width=0.13\textwidth]{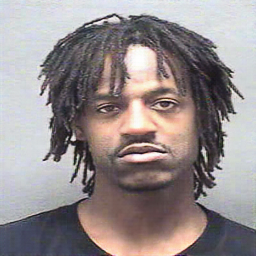} &
            \includegraphics[width=0.13\textwidth]{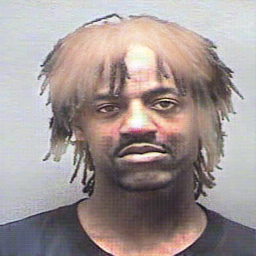} &&
            \includegraphics[width=0.13\textwidth]{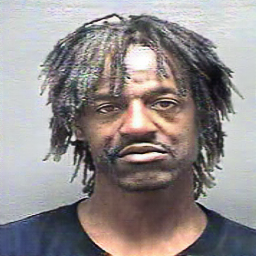} &
            \includegraphics[width=0.13\textwidth]{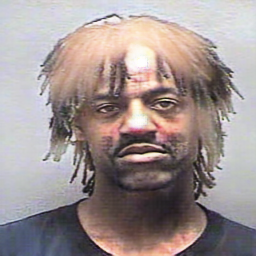} &&
            \includegraphics[width=0.13\textwidth]{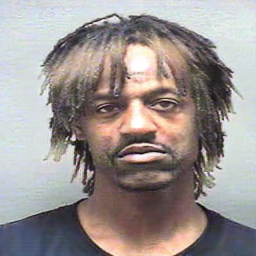} \\
                25  && \red{32 (-1)} & 37 (+4) && \red{47 (-1)} & 38 (-10) && 37 (-11)\\
            
            \includegraphics[width=0.13\textwidth]{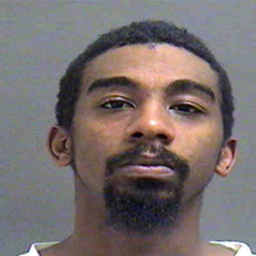} &&
            \includegraphics[width=0.13\textwidth]{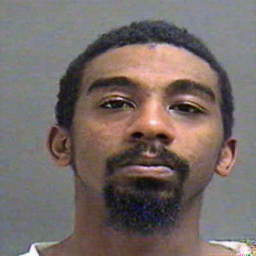} &
            \includegraphics[width=0.13\textwidth]{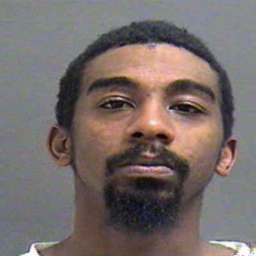} &&
            \includegraphics[width=0.13\textwidth]{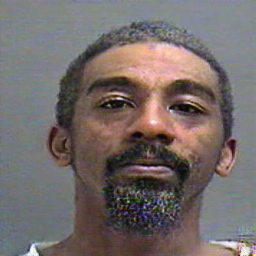} &
            \includegraphics[width=0.13\textwidth]{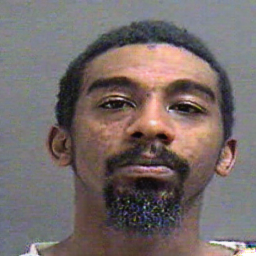} &&
            \includegraphics[width=0.13\textwidth]{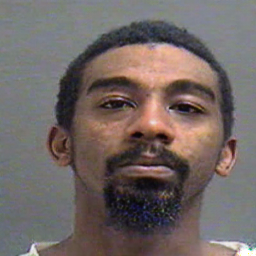} \\
                28  && \red{32 (-1)} & 39 (+6) && \red{54 (+6)} & 42 (-6) && 42 (-6)\\
            
            \includegraphics[width=0.13\textwidth]{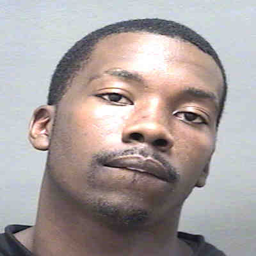} &&
            \includegraphics[width=0.13\textwidth]{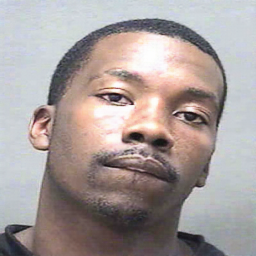} &
            \includegraphics[width=0.13\textwidth]{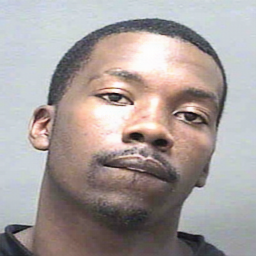} &&
            \includegraphics[width=0.13\textwidth]{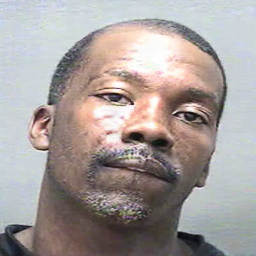} &
            \includegraphics[width=0.13\textwidth]{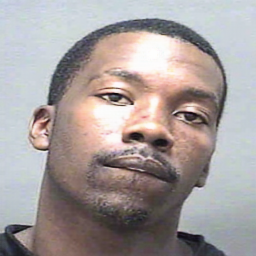} &&
            \includegraphics[width=0.13\textwidth]{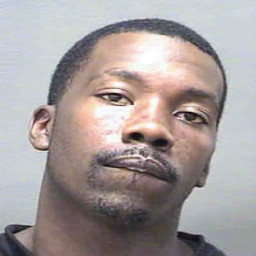} \\
                25  && \red{28 (-5)} & 28 (-5) && \red{51 (+3)} & 36 (-12) && 40 (-8)\\
            
            Input && BiTrackGAN & Group-GAN && BiTrackGAN & Group-GAN && CycleGAN \\
            
        \end{tabular}
    
    \caption{Facial image aging synthesis comparison. BiTrackGAN, Group-GAN, and CycleGAN are utilized to synthesize facial image aging for some subjects. For each method, facial images from the age group 18-23 are translated into the target age group 45-52 and intermediate state age group 30 to 35. A Face++ evaluated age is displayed under each subject's facial image.}
    \label{fig:Aging_Compared_With_3_Methods}
\end{figure*}

\begin{figure*}[htbp]
    \centering
        \begin{tabular}{cccccccccc}
            \multicolumn{9}{c}{$\overbrace{\rule{41.5em}{0pt}}^{\text{Face Rejuvenation}}$}\\
            Age 45-52 && \multicolumn{2}{c}{Age 30-35} && \multicolumn{2}{c}{Age 18-23} && Age 18-23\\
            Median=48 && \multicolumn{2}{c}{Medium=33} && \multicolumn{2}{c}{Median=20} && Median=20\\
            
            \includegraphics[width=0.13\textwidth]{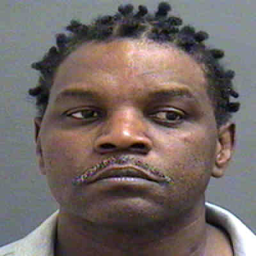} &&
            \includegraphics[width=0.13\textwidth]{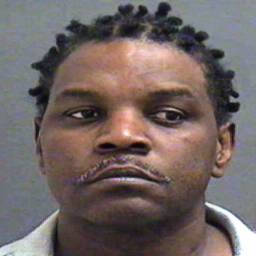} &
            \includegraphics[width=0.13\textwidth]{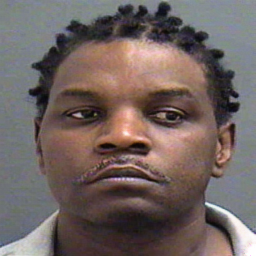} &&
            \includegraphics[width=0.13\textwidth]{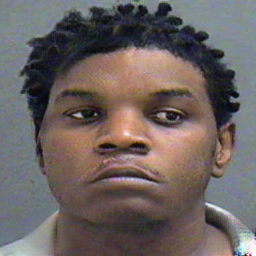} &
            \includegraphics[width=0.13\textwidth]{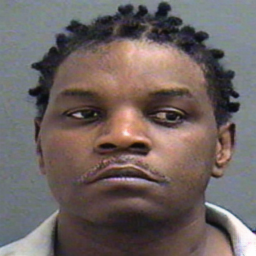} &&
            \includegraphics[width=0.13\textwidth]{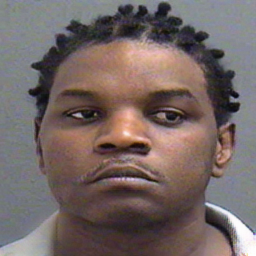} \\
                39  && \red{35 (+2)} & 30 (-3) && \red{20 (+0)} & 24 (+4) && 24 (+4) \\
                  
            \includegraphics[width=0.13\textwidth]{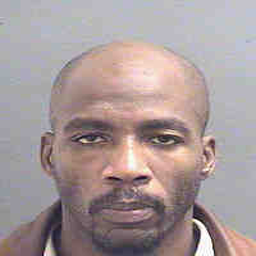} &&
            \includegraphics[width=0.13\textwidth]{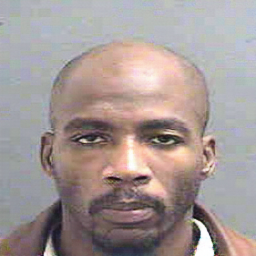} &
            \includegraphics[width=0.13\textwidth]{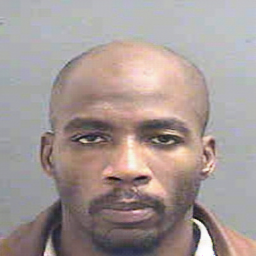} &&
            \includegraphics[width=0.13\textwidth]{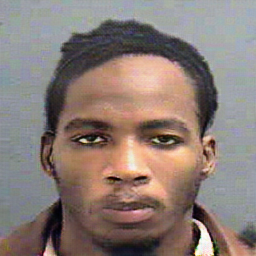} &
            \includegraphics[width=0.13\textwidth]{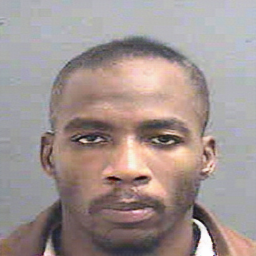} &&
            \includegraphics[width=0.13\textwidth]{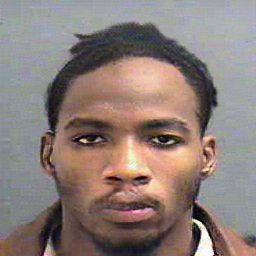} \\
                41  && \red{37 (+4)} & 37 (+4) && \red{22 (+2)} & 32 (+12) && 21 (+1) \\
                  
            \includegraphics[width=0.13\textwidth]{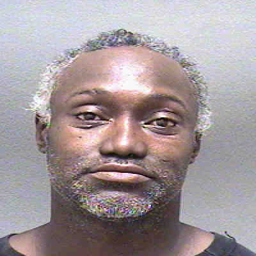} &&
            \includegraphics[width=0.13\textwidth]{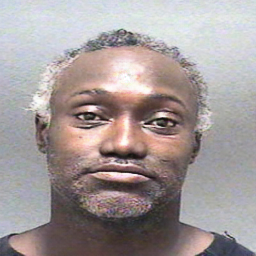} &
            \includegraphics[width=0.13\textwidth]{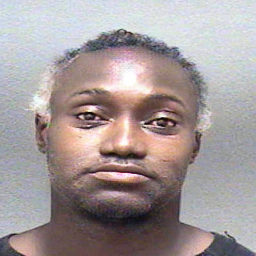} &&
            \includegraphics[width=0.13\textwidth]{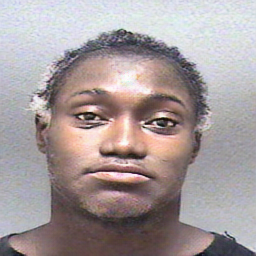} &
            \includegraphics[width=0.13\textwidth]{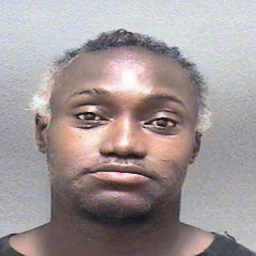} &&
            \includegraphics[width=0.13\textwidth]{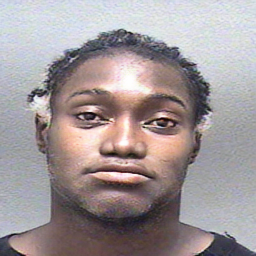} \\
                43  && \red{35 (+2)} & 27 (-6) && \red{21 (+1)} & 24 (+4) && 21 (+1) \\
                  
            \includegraphics[width=0.13\textwidth]{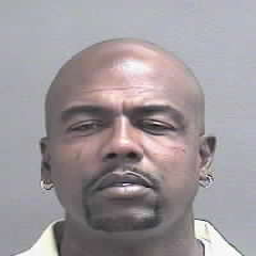} &&
            \includegraphics[width=0.13\textwidth]{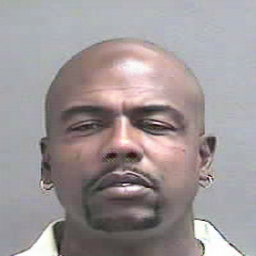} &
            \includegraphics[width=0.13\textwidth]{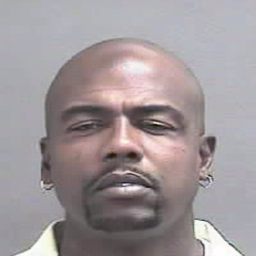} &&
            \includegraphics[width=0.13\textwidth]{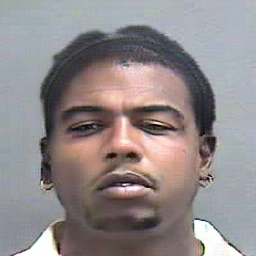} &
            \includegraphics[width=0.13\textwidth]{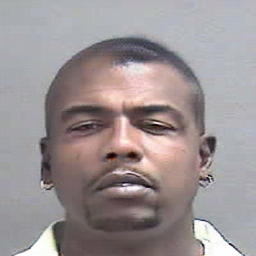} &&
            \includegraphics[width=0.13\textwidth]{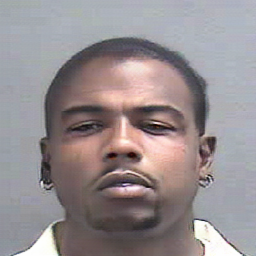} \\
                42  && \red{38 (+5)} & 39 (+6) && \red{23 (+3)} & 37 (+17) && 22 (+2) \\
                  
            \includegraphics[width=0.13\textwidth]{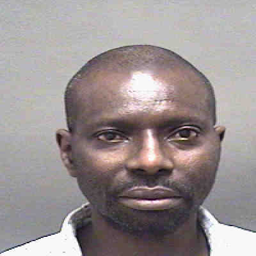} &&
            \includegraphics[width=0.13\textwidth]{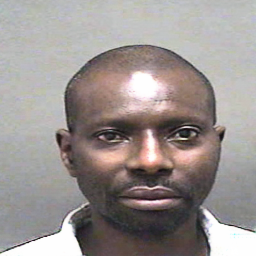} &
            \includegraphics[width=0.13\textwidth]{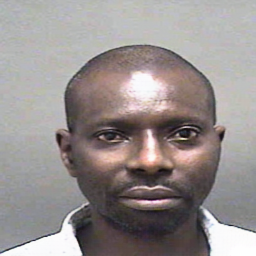} &&
            \includegraphics[width=0.13\textwidth]{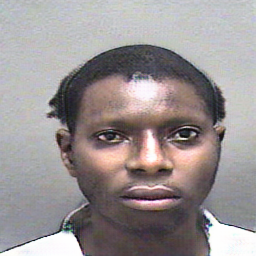} &
            \includegraphics[width=0.13\textwidth]{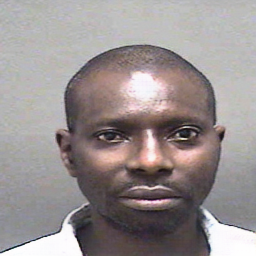} &&
            \includegraphics[width=0.13\textwidth]{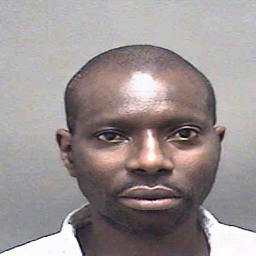} \\
                40  && \red{35 (+2)} & 35 (+2)  && \red{22 (+2)} & 29 (+9) && 31 (+11)  \\
            
            Input && BiTrackGAN & Group-GAN && BiTrackGAN & Group-GAN && CycleGAN \\
            
        \end{tabular}
    
    \caption{Facial image rejuvenation synthesis comparison. BiTrackGAN, Group-GAN, and CycleGAN are utilized to synthesize facial image rejuvenation for some subjects. For each method, facial images from the age group 45-52 are translated into the target age group 18-23 and intermediate state age group 30 to 35. A Face++ evaluated age is displayed under each subject's facial image.}
    \label{fig:Rejuvenation_Compared_With_3_Methods}
\end{figure*}

\begin{figure*}[htbp]
    \includegraphics[width = 1.00\textwidth]{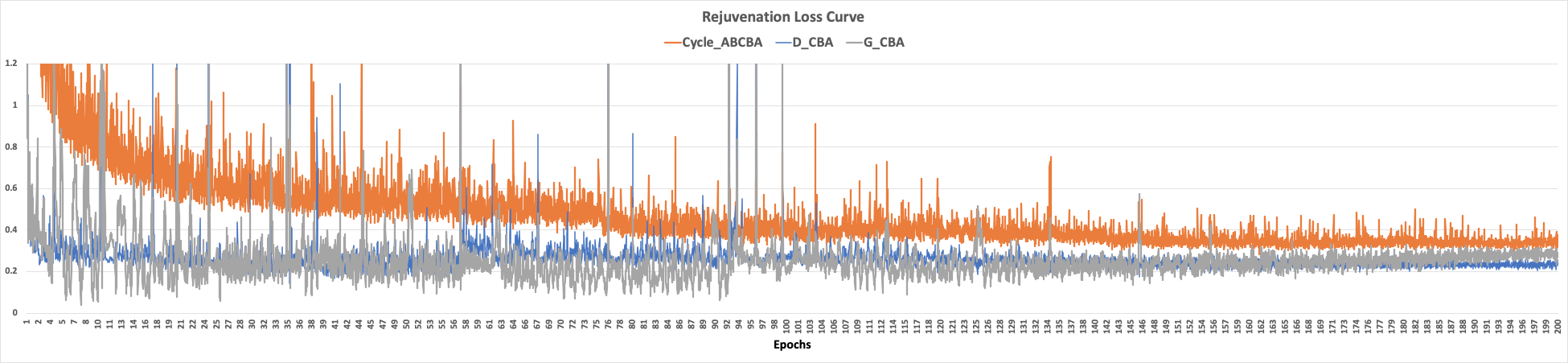}\par
    \caption{Rejuvenation loss graph. The age groups 18-23, 30-35, and 45-52 are represented by A, B, and C. Cycle\_ABCBA represents the translation loss from A to B to C to B to A in the BiTrackGAN forward cycle training. D\_CBA represents the discriminator loss of translating from C to B to A in BiTrackGAN training. G\_CBA represents the generator loss of translating from C to B to A in BiTrackGAN training.}
    \label{fig:Rejuvenation_Loss_D_CBA_cycle_ABCBA_G_CBA}
    \centering
    
    \includegraphics[width = 1.00\textwidth]{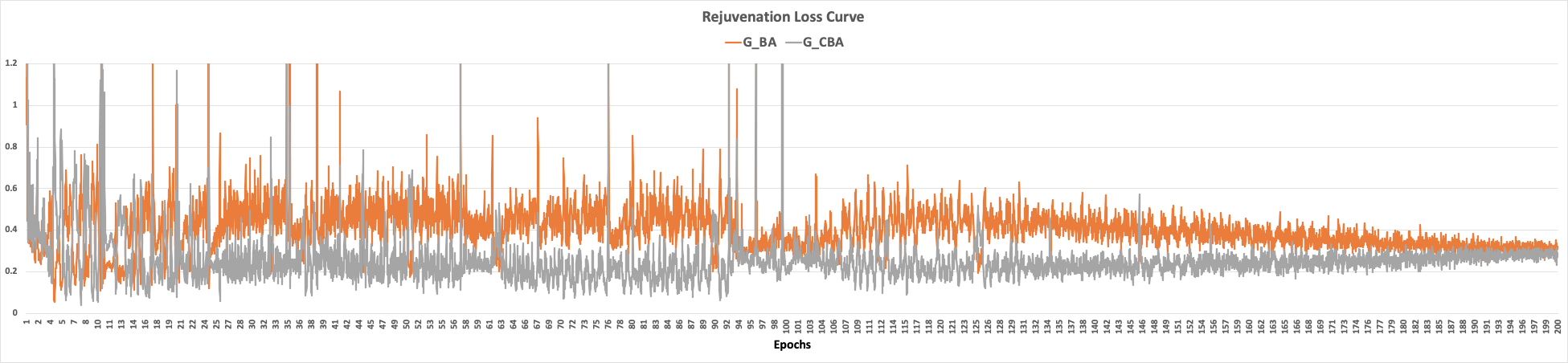}\par
    \caption{G\_CBA represents the generator loss of translating from C to B to A in BiTrackGAN training. G\_BA represents the generator loss of translating from B to A in CycleGAN\_1 training.}\par
    \label{fig:Rejuvenation_Loss_G_CB_G_BA_G_CBA}
    \centering
\end{figure*}

\begin{figure*}[htbp]
    \includegraphics[width = 1.00\textwidth]{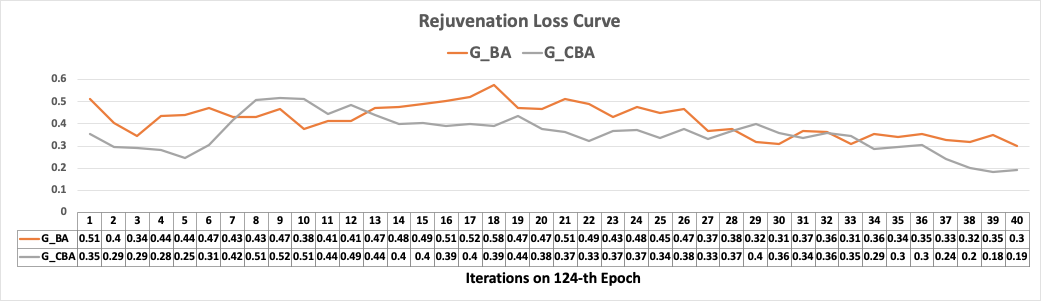}\par 
    \centering

    \includegraphics[width = 1.00\textwidth]{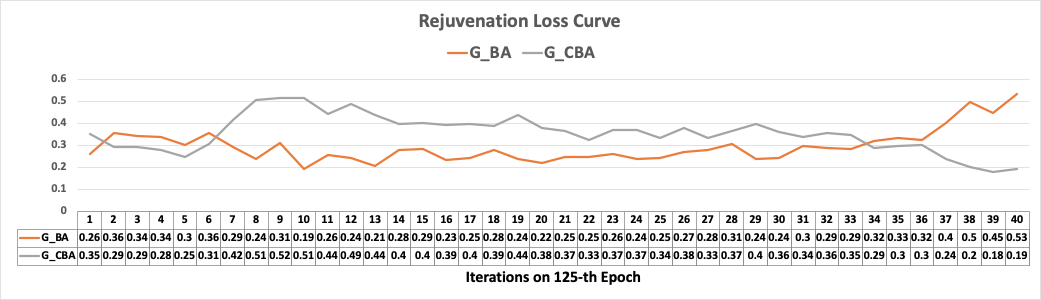}\par
    \centering

\caption{At epochs 124 and 125, loss of G\_BA and G\_CBA show a seesaw phenomenon before steady progressive convergence. In other words, when G\_BA loss decreases, G\_CBA loss increases, and vice versa.}
\label{fig:Rejuvenatin_Loss_G_BA_G_CBA_124_125_epoch}
\end{figure*}

\subsection{Qualitative Comparison}

In this subsection, we first compare cross-age synthesis performance between our work, Group-GAN, and CycleGAN. For a fair assessment of whether the synthesis image is closer to the target age, for instance, 33 in the 30-35 age group, we leverage the Face++ online API, which is widely used to detect face attributes. Then, by observing the training loss curve, we discuss how the bottom-up training approach induces a constraint mechanism to achieve smoother, progressive synthesis and diversity. 

\subsubsection{Reasonable Aging Effect with Constraint Mechanism}

In Figure~\ref{fig:Aging_Compared_With_3_Methods}, our work achieves a more reasonable synthesis effect for the 45-52 age group than others. Similarly, Figure~\ref{fig:Rejuvenation_Compared_With_3_Methods} also shows a good performance regarding rejuvenation

In the 30-35 age group, as shown in Figure~\ref{fig:Aging_Compared_With_3_Methods} and Figure~\ref{fig:Rejuvenation_Compared_With_3_Methods}, our work performs closer to the target, namely the age of 33. Age groups 30 to 35 play an important role in intermediate states; whatever the direction of translation, it must pass through the intermediate state in an entire translation pipeline, namely the constraint generation mechanism. As we can expect, a smoother, more progressive final synthesis will be achieved if the intermediate state is ideal to approach the target distribution. There is no doubt that a reasonable aging effect is more important than a stronger one. Our work achieve the target age better than others in the age groups 45-52 in Figure~\ref{fig:Aging_Compared_With_3_Methods}, or 18-23 in Figure~\ref{fig:Rejuvenation_Compared_With_3_Methods}.

\subsubsection{Diversity}

In the GAN approach, mode collapse is a well-known issue. It may result in fewer patterns in the synthesis than it should be. Again, in Figure~\ref{fig:Rejuvenation_Compared_With_3_Methods}, hairstyles in our work are more diverse than other works in the target age group 18-23. Based on the youth hairstyle in the dataset, the rejuvenation effect appears more reasonable for youth aged 18-23.

\subsubsection{Analysis of Loss Curve}

In BiTrackGAN, the patterns of training loss in aging and rejuvenation are similar, so we can just discuss the trend in rejuvenation.

Firstly, we observe G\_CBA loss, Cycle\_ABCBA loss, and D\_CBA loss, as they are related. D\_CBA and Cycle\_ABCBA are shown to converge steadily in Figure~\ref{fig:Rejuvenation_Loss_D_CBA_cycle_ABCBA_G_CBA}, while G\_CBA oscillates widely early in training. 

Secondly, in Figure~\ref{fig:Rejuvenation_Loss_G_CB_G_BA_G_CBA}, the loss of G\_CBA in BiTrackGAN and the loss of G\_BA in CycleGAN 1 converge gradually to 0.2-0.4, which is logical since both have the same objective and use the same discriminator. 

Finally, we find that loss of G\_BA and G\_CBA present a seesaw phenomenon before progressive steady convergence, such as at epochs 124 and 125 as shown in Figure~\ref{fig:Rejuvenatin_Loss_G_BA_G_CBA_124_125_epoch}. In other words, when G\_BA loss decreases, G\_CBA loss increases, and vice versa.

To explain this phenomenon, we can recall that BiTrackGAN is trained from the bottom up. In this approach, for each iteration in training, the training order is CycleGAN 1, CycleGAN 2, and BiTrackGAN, with CycleGAN 1 translating between age groups A and B, CycleGAN 2 translating between age groups B and C, and BiTrackGAN translating between age groups A, B, and C. 

Since the G\_CBA in BiTracGAN cascades through the G\_CB in CycleGAN 2, and the G\_BA in CycleGAN 1, optimization of the G\_CBA may result in a slight increase in the G\_CB and G\_BA losses. As a result, they are forced to re-explore better distribution to approach the target. Consequently, with a reasonable output for age group B, age group A will have a reasonable output as well.

Overall, with the bottom-up training approach, each generator has more opportunities to re-explore the better distribution, which enables it to synthesize a more reasonable output for the next cascading generator. It is our claim that bottom-up training can introduce constraint mechanisms in the translation pipeline that enable a more reasonable and diverse output.

\subsection{Quantitative Analysis}
\begin{figure}[htbp]
    \centering
    \includegraphics[width=0.46\textwidth]{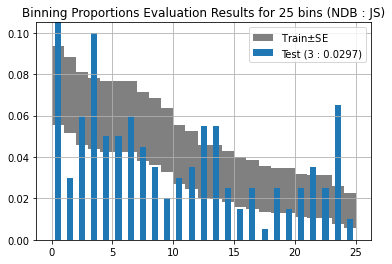}
    \caption{In this NDB test, K=25 was used to assess the rejuvenation of BiTrackGAN to 18-23 years of age.}
    \label{fig:NDB_BiTrackGAN_FakeBA_K_25}
\end{figure}

\begin{table}[htbp]
    \begin{subtable}[h]{0.45\textwidth}
        \centering
        \begin{tabular}{l | l | l | l| l}
                            &NDB/K(MEAN)    &NDB/K(SD)      &JS(MEAN)   &JS(SD) \\
        \hline \hline
            BiTrackGAN      &0.0500         &0.1100         &0.0030     &0.0023     \\
            Group-GAN       &0.6300         &0.0732         &0.0364     &0.0024     \\
            CycleGAN        &0.3800         &0.0616         &0.0093     &0.0012     \\
       \end{tabular}
       \caption{K=5}
       \label{tab:k=5}
    \end{subtable}

    \begin{subtable}[h]{0.45\textwidth}
        \centering
        \begin{tabular}{l | l | l | l| l}
                            &NDB/K(MEAN)    &NDB/K(SD)      &JS(MEAN)   &JS(SD)     \\
        \hline \hline
            BiTrackGAN      &0.1550         &0.0760         &0.0132     &0.0040     \\
            Group-GAN       &0.6000         &0.0649         &0.0730     &0.0050     \\
            CycleGAN        &0.3600         &0.0940         &0.0282     &0.0042     \\
       \end{tabular}
       \caption{K=10}
       \label{tab:k=10}
    \end{subtable}

    \begin{subtable}[h]{0.45\textwidth}
        \centering
        \begin{tabular}{l | l | l | l| l}
                            &NDB/K(MEAN)    &NDB/K(SD)      &JS(MEAN)   &JS(SD)     \\
        \hline \hline
            BiTrackGAN      &0.2333         &0.1049         &0.0251     &0.0052     \\
            Group-GAN       &0.4900         &0.0899         &0.0806     &0.0080     \\
            CycleGAN        &0.2800         &0.0634         &0.0359     &0.0044     \\
       \end{tabular}
       \caption{K=15}
       \label{tab:k=15}
    \end{subtable}

    \begin{subtable}[h]{0.45\textwidth}
        \centering
        \begin{tabular}{l | l | l | l| l}
                            &NDB/K(MEAN)    &NDB/K(SD)      &JS(MEAN)   &JS(SD)     \\
        \hline \hline
            BiTrackGAN      &0.1760         &0.0586         &0.0345     &0.0050     \\
            Group-GAN       &0.4260         &0.0690         &0.0980     &0.0067     \\
            CycleGAN        &0.2180         &0.0543         &0.0480     &0.0056     \\
       \end{tabular}
       \caption{K=25}
       \label{tab:k=25}
    \end{subtable}
    
    \begin{subtable}[h]{0.45\textwidth}
        \centering
        \begin{tabular}{l | l | l | l| l}
                            &NDB/K(MEAN)    &NDB/K(SD)      &JS(MEAN)   &JS(SD)     \\
        \hline \hline
            BiTrackGAN      &0.1000         &0.0337         &0.0586     &0.0082     \\
            Group-GAN       &0.1690         &0.0334         &0.1248     &0.0085     \\
            CycleGAN        &0.1180         &0.0337         &0.0748     &0.0069     \\
       \end{tabular}
       \caption{K=50}
       \label{tab:k=50}
    \end{subtable}
    
    
    \caption{Bin-proportional NDB/K scores for different models trained on MORPH II, using 4000 samples from the training set and 200 samples generated from each model, for different numbers of bins (K). The listed values are NDB numbers of statistically different bins, with a significant level of 0.05, divided by the number of bins K (lower is better).}
    \label{tab:NDB_Score_Various_K}
\end{table}

To identify the degree of mode collapse, we conduct NDB testing with K=50, 25, 15, 10, and 5. For each K, these tests are performed 20 times to calculate the statistical mean and standard deviation of NDB/K and Jensen Shannon Divergence (JS). As shown in Table~\ref{tab:NDB_Score_Various_K}, BiTrackGAN clearly outperforms other works. As Figure~\ref{fig:NDB_BiTrackGAN_FakeBA_K_25}, we also show the binning proportion evaluation for NDB test with K=25. 

Additionally, small K implies that samples are divided into fewer classes based on similar features. It is likely that these similar features represent certain facial attributes, such as hairstyles or beard styles, in facial images.


Based on the above comparison analysis, we can summarize as follows:
\begin{itemize}
    \item The constraint mechanism in cross-age aging synthesis, namely the ideal intermediate state, plays a crucial role. 
    \item The bottom-up training strengthens the aging effect between small difference domains and synthesizes more reasonable and diverse aging.
    \item It should be a good design to build BiTrackGAN as a module so that more extensive applications can be built upon it.
\end{itemize}
While BiTrackGAN produces excellent aging results, it requires a lot of neural network parameters, and powerful parallel accelerating computing resources. This can be an interesting performance-related research topic in the future. 
\section{Conclusion}
\label{V_Conclusion}

This paper proposes a network named BiTrackGAN, which adopts a bottom-up training manner for two cascaded CycleGAN blocks to synthesize facial aging and rejuvenation. It induces a constraint mechanism between these two CycleGAN blocks in this training manner.

According to our qualitative comparison in experiments, our work results in a more reasonable facial aging and rejuvenation effect. Moreover, the NDB testing has shown that our proposed has a lower mode collapse degree in quantitative analysis.

BiTrackGAN achieves a more reasonable and accurate synthesis based on the CycleGAN approach by inducing an intermediate state as a constraint mechanism. In an accuracy-intensive field, such as tumor growth prediction in medical images, BiTrackGAN has the potential to achieve more excellent results.

In addition, BiTrackGAN can be used as a module to build more extensive applications.

\bibliographystyle{IEEEtran} 
\bibliography{./bib/ref.bib}

\end{document}